\documentclass[10pt,twocolumn,letterpaper]{article}

\usepackage{cvpr}
\usepackage{times}
\usepackage{epsfig}
\usepackage{graphicx}
\usepackage{amsmath}
\usepackage{amssymb}
\usepackage{booktabs}
\usepackage{subcaption}
\usepackage{authblk}
\usepackage{microtype}

\usepackage[pagebackref=true,breaklinks=true,letterpaper=true,colorlinks,bookmarks=false]{hyperref}

\cvprfinalcopy 

\def\cvprPaperID{649} 

\ifcvprfinal\fi
\begin{document}

\newcommand{\multilinecell}[2][c]{%
  \begin{tabular}[#1]{@{}c@{}}#2\end{tabular}}
  
\title{Pointwise Convolutional Neural Networks}

\author[ ]{Binh-Son Hua}
\author[ ]{Minh-Khoi Tran}
\author[ ]{Sai-Kit Yeung}
\affil[ ]{The University of Tokyo \quad \quad \quad \quad Singapore University of Technology and Design}
\renewcommand\Authsep{\hspace{0.8in}}
\renewcommand\Authand{\hspace{0.8in}}
\renewcommand\Authands{\hspace{0.8in}}

\maketitle
\thispagestyle{empty}
 
\footnotetext{This work was done when Binh-Son Hua was a postdoctoral researcher in Singapore University of Technology and Design in 2017.}

\begin{abstract}
   Deep learning with 3D data such as reconstructed point clouds and CAD models has received great research interests recently. However, the capability of using point clouds with convolutional neural network has been so far not fully explored. 
   In this paper, we present a convolutional neural network for semantic segmentation and object recognition with 3D point clouds. At the core of our network is \emph{pointwise convolution}, a new convolution operator that can be applied at each point of a point cloud. Our fully convolutional network design, while being surprisingly simple to implement, can yield competitive accuracy in both semantic segmentation and object recognition task.
\end{abstract}

\section{Introduction}

Deep learning with 3D data has received great research interests recently, which leads to noticeable advances in typical applications including scene understanding, shape completion, and shape matching. Among these, scene understanding is considered as one of the most important tasks for robots and drones as it can assist exploratory scene navigations. Tasks such as semantic scene segmentation and object recognition are often performed to predict contextual information about objects for both indoor and outdoor scenes. 

\begin{figure}[t]
	\centering
	\includegraphics[width=0.9\linewidth]{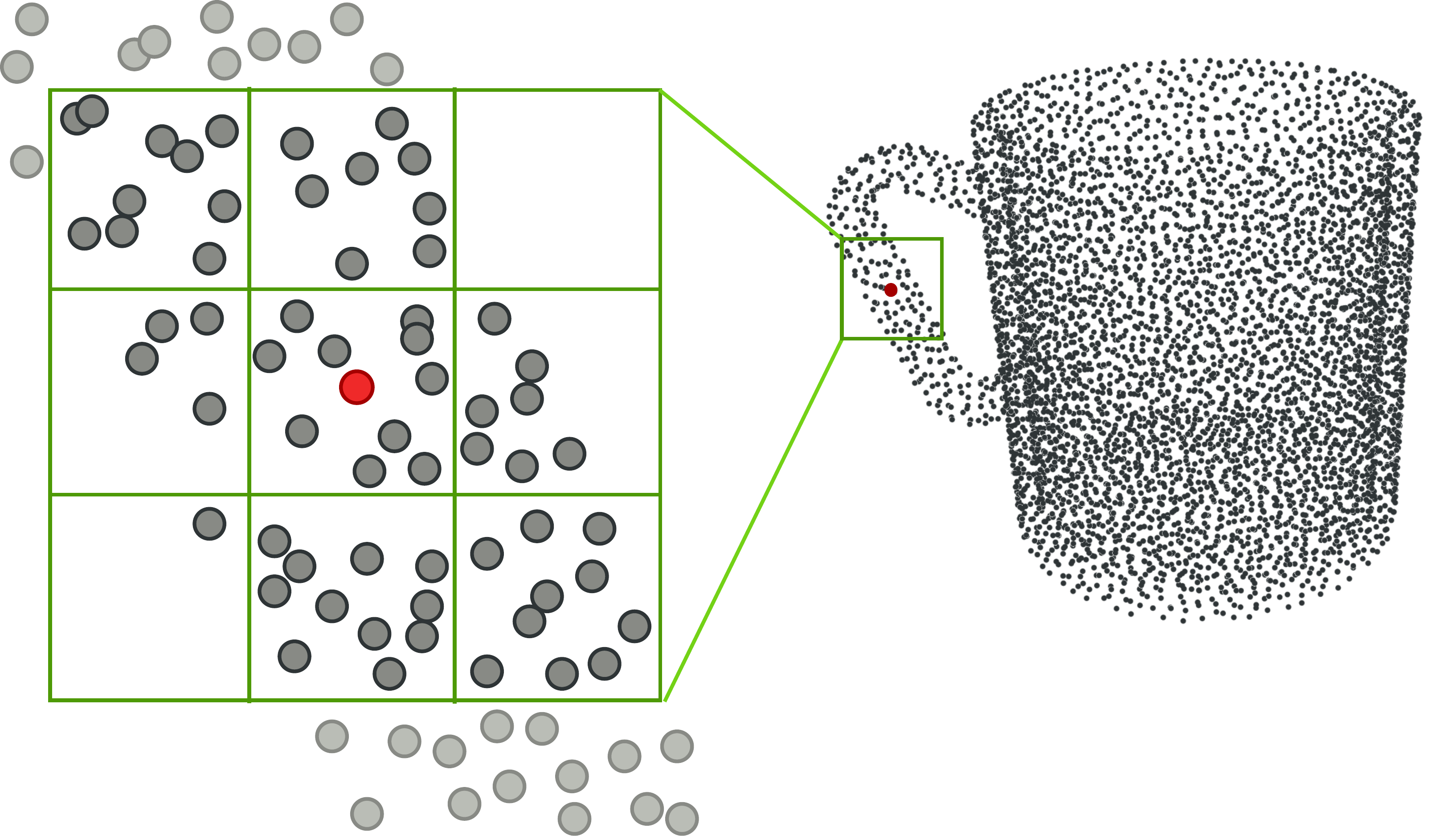}
	\caption{Pointwise convolution. We define a new convolution operator for point cloud input. For each point, nearest neighbors are queried on the fly and binned into kernel cells before convolving with kernel weights. By stacking pointwise convolution operators together, we can build fully convolutional neural networks for scene segmentation and object recognition for point clouds.}
	\label{fig:teaser}
\end{figure}

Unfortunately, deep learning in 3D was deemed difficult due to the fact that there are several ways to represent 3D data such as volumes, point clouds, or multi-view images. Volume representation is a true 3D representation and straightforward to implement but often requires a large amount of memory for data storage. By contrast, multi-view representation is not a true 3D representation but shows promising prediction accuracy as existing pre-trained weights from 2D networks can be utilized. Among such representations, point clouds have been the most flexible as they are compact and could be exported from a wide range of CAD modelling and 3D reconstruction software. However, the capability of using point clouds with neural network has been so far not fully explored. 

In this paper, we present a convolutional neural network for semantic segmentation and object recognition with 3D point clouds. At the core of our network is a new convolution operator, called \emph{pointwise convolution}, which can be applied at each point in a point cloud to learn pointwise features. This leads to surprisingly simple and fully convolutional network designs for scene segmentation and object recognition. Our experiments show that pointwise convolution can yield competitive accuracy to previous techniques while being much simpler to implement. In summary, our contributions are:
\begin{itemize}
	\item A pointwise convolution operator that can output features at each point in a point cloud;
	\item Two pointwise convolutional neural networks for semantic scene segmentation and object recognition.
\end{itemize}

\begin{figure*}[t]  
	\includegraphics[width=\linewidth]{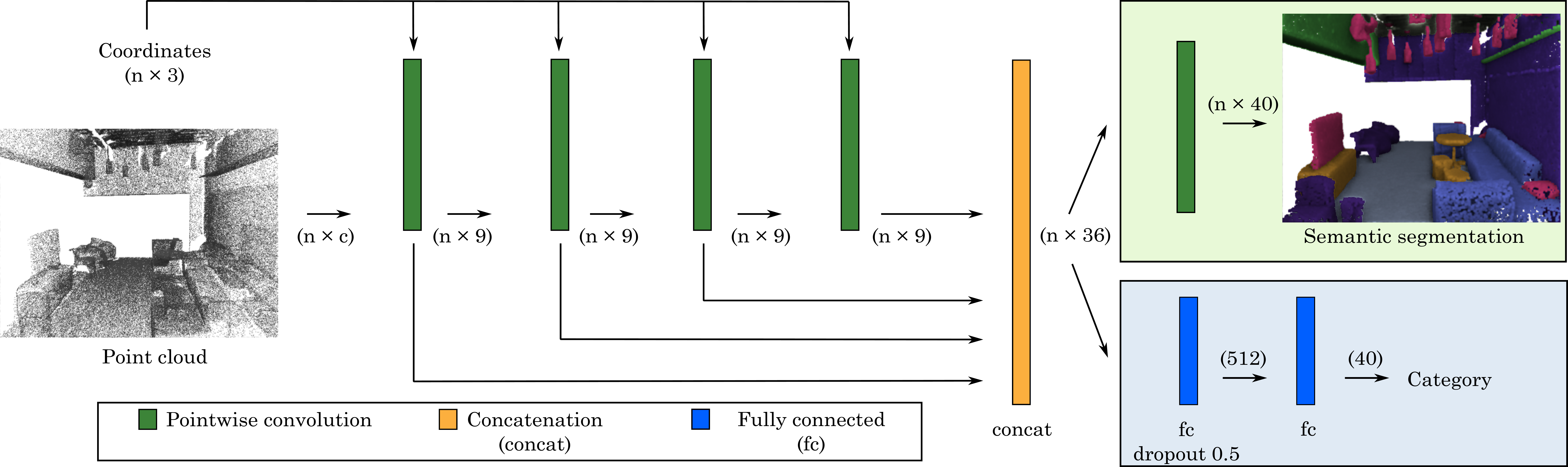}
	\caption{Pointwise convolutional neural network. The input point cloud is fed into each convolution operator, and all outputs are concatenated before being fed to a a final convolution layer for dense semantic segmentation, or to fully connected layers for object recognition. In this figure, we assume point cloud with $n$ points and $c$ attributes (colors, normals, coordinates, etc.). We use 9 output channels for each convolution operator before concatenation. Source code is available at our homepage~\cite{hua-scenenn-3dv16}.}
	\label{fig:network}
\end{figure*}
\section{Related Works}
Recently, there has been a great number of works about deep learning with 3D data. Let us focus on those for scene understanding tasks such as semantic segmentation and object recognition. 

\subsection{Shape descriptors}
Hand-crafted shape descriptors were widely used in computer vision and graphics applications before the era of deep learning. For example, 3D shapes can be projected into 2D images and represented by a set of 2D descriptors on such images. Shapes can then be represented as histograms or bag-of-feature models which can be constructed from surface normals and curvatures \cite{horn-extended-1984}. 
3D shapes can also be represented by their inherent statistical properties, such as distance distribution \cite{osada2002shape} and harmonic descriptors \cite{kazhdan-rotation-2003}. Heat kernel signatures extract shape descriptions by simulating an heat diffusion process on 3D shapes \cite{sun-concise-2009}. The Light Field Descriptor (LFD) is another popular descriptor useful in the shape classification tasks. It extracts geometric and Fourier descriptors from object silhouettes rendered from several different viewpoints \cite{chen2003visual}.
Despite their long history and being widely used, hand-crafted 3D shape descriptors do not generalize well across different domains. 

\subsection{Object recognition}
Convolutional neural networks (CNNs) \cite{lecun1998gradient} has been successfully applied in various areas of computer vision and artificial intelligence. Recently, significant achievements have been reached in understanding images through learning features by CNNs. Large RGB image datasets like ImageNet \cite{deng2009imagenet} can be used in training a CNN, which is in turn able to learn general purpose image descriptors from such datasets. Image descriptors generated by CNNs are proved to greatly outperform other hand-crafted features for various tasks, including object detection \cite{girshick2014rich}, scene recognition \cite{donahue2014decaf}, texture recognition \cite{sharif2014cnn,cimpoi2014describing} and classification \cite{han2015matchnet}.

Recently, several approaches to using 3D convolutional networks to extract shape descriptor have been proposed, ranging from voxel-based representation~ \cite{wu-3dshapenets-cvpr15,maturana-voxnet-iros15}  panorama~\cite{shi2015deeppano}, feature pooling from 2D projections from multiple viewpoints~\cite{su15mvcnn,qi-volumetric-cvpr16}, to point set~\cite{qi-volumetric-cvpr16}. 
Among these, Qi \etal~\cite{qi-pointnet-cvpr17}'s PointNet is one of the first network architectures that can handle point cloud data. PointNet is robust as it can learn an order-invariance function to canonicalize input point clouds. Subsequently, PointCNN~\cite{li-pointcnn-ar18} explored the idea of equivariance instead of invariance and demonstrated competitive performance to PointNet. To achieve scalability, it is also possible to learn representations on unstructured point clouds by building computational graphs based on hierarchical data structures such as octree~\cite{riegler-OctNet-cvpr17} and kd-tree~\cite{klokov-escape-iccv17}. 

Despite their competitive performance, network structures based on PointNet~\cite{qi-pointnet-cvpr17} are rather complex.
%
%
In this work, we show that it is possible to perform scene understanding tasks such as semantic segmentation and object recognition on \emph{ordered} point clouds. We design pointwise convolution, a simple convolution operator for 3D point cloud and use it to make (fully) convolutional neural networks for object recognition and semantic segmentation. With the availability of our pointwise convolution, we aim to pave the way towards adapting many existing network architecture designed for scene understanding with color and RGB-D images~\cite{simonyan-vgg-ar14,long-fcn-cvpr15,song-sunrgbd-cvpr15} to the 3D domain.

%

\subsection{Semantic segmentation}
There are considerably great numbers of related works in  semantic segmentation. 
Since the introduction of the NYUv2 dataset from Silberman~\etal~\cite{nathan-nyu-eccv12},
there has been a spark in the direction of RGBD semantic segmentation. The work from 
Long~\etal~\cite{long-fcn-cvpr15} showed how to adopt a conventional classifcation
network for the semantic segmentation problem. Since then, different techniques have
been proposed to further improve the segmentation results. Some notable examples are
SegNet~\cite{badrinarayanan-segnet-arxiv15} which employs an encoder-decoder architecture,
or the dilation filter~\cite{yu-dilated-iclr16}.

In the 3D domain, interactive semantic segmentation~\cite{valentin-semanticpaint-tog15,anno-tvcg17} relied on user strokes 
to propagate segmentation. McCormac \etal~\cite{handa-semantic-ar16} explored 
transfering semantic segmentation from 2D predictions to the 3D domain. An advantage 
of such methods is that they can produce high-resolution segmentation. However, none of 
the predictions can be performed directly in the 3D domain.

SSCNet~\cite{song-sscnet-cvpr17} applied convolutional neural network to a 3D volume 
representation to classify each voxel in the scene. This could be flexible as real-time 
scene reconstruction techniques such as KinectFusion~\cite{newcombe-kinect-ismar11} and 
voxel hashing~\cite{niessner-hash-tog13} are often based on volumes. 
PointNet~\cite{qi-pointnet-cvpr17} can also be used for semantic segmentation with 
minor modifications from their object recognition network. 

Recently, Qi \etal~\cite{qi-3dgraph-iccv17} proposed to build a graph neural network for semantic segmentation on a point cloud, where each graph node is a group of points and graph edges are constructed by nearest neighbor search on the point cloud. Their results are shown with RGB-D images, where color features from a pre-trained VGG-16 network~\cite{simonyan-vgg-ar14,chen-deeplab-ar14} are used to initialize the prediction. 
Here, we demonstrate a fully convolutional neural network for 3D point cloud segmentation. Compared to the method by Qi \etal~\cite{qi-3dgraph-iccv17}, we train our network from scratch. The input point cloud is also more general such as CAD models or 3D meshes reconstructed from RGB-D sensors.

\section{Pointwise Convolution}
Before presenting pointwise convolution, we briefly revise a few possibilities to represent 3D data for neural network. The most straightforward approach is perhaps to employ volumetric representation. For example, VoxNet~\cite{maturana-voxnet-iros15} represents each object by a volume up to $64 \times 64 \times 64$ resolution. This is natural because almost existing network architecture for image applications can be adopted. However, a significant drawback is that volumetric representation requires a large amount of memory while the number of non-zero values in a volume only accounts for a very small percentage. This could be addressed by a sparse representation~\cite{riegler-OctNet-cvpr17}. 

A second possibility is to use point clouds. This is a direct representation as point cloud is often the output of many applications such as RGB-D reconstruction and CAD modeling. However, mapping point cloud to neural network is not natural because traditional convolution operators are only designed for grid and volumes. PointNet~\cite{qi-pointnet-cvpr17} implements point feature learning by fully connected layers. 

The previous limitations motivate us to design fully convolutional networks for point clouds.
The basic building block of our architecture is a convolution operator applied at each point in a point cloud, which we term the \emph{pointwise convolution}. This operator works as follows. 

\paragraph{Convolution.}
A convolution kernel is centered at each point of a point cloud. Neighbor points within the kernel support can contribute to the center point. Each kernel has a size or radius value, which can be adjusted to account for different number of neighbor points in each convolution layer. Figure~\ref{fig:teaser} shows a diagram that demonstrates this idea.

Formally, pointwise convolution can be written as
\begin{align}
x_i^\ell = \sum_{k} w_k \frac{1}{\mid \Omega_i(k) \mid} \sum_{p_j \in \Omega_i(k)} x_j^{\ell - 1},
\end{align}
where $k$ iterates over all sub-domains in the kernel support; $\Omega_i(k)$ is the $k$-th sub-domain of the kernel centered at point $i$; $p_i$ is the coordinate of point $i$; $\mid \cdot \mid$ counts all points within the sub-domain; $w_k$ is the kernel weight at the $k$-th sub-domain, $x_i$ and $x_j$ the value at point $i$ and $j$, and $\ell - 1$ and $\ell$ the index of the input and output layer. 

\paragraph{Gradient backpropagation.} To make pointwise convolution trainable, it is necessary to compute the gradients with respects to the input data and the kernel weights. Let $L$ is the loss function. The gradient with respect to input could be defined as
\begin{align}
\frac{ \partial L }{\partial x_j^{\ell - 1}} = \sum_{i \in \Omega_j} \frac{ \partial L }{\partial x_i^{\ell}} \frac{ \partial x_i^\ell }{\partial x_j^{\ell - 1}} 
\end{align}
where we iterate over all neighbor points $i$ of a given point $j$. In the chain rule, ${ \partial L } / {\partial x_i^{\ell}}$ is the gradient up to layer $\ell$, which is known during back propagation. The derivative ${ \partial x_i^\ell } / {\partial x_j^{\ell - 1}}$ could be written as
\begin{align}
\frac{ \partial x_i^\ell }{\partial x_j^{\ell - 1}} = \sum_{k} w_k \frac{1}{\mid \Omega_i(k) \mid} \sum_{p_j \in \Omega_i(k)} 1
\end{align}

Similarly, the gradient with respect to kernel weights could be defined by iterating over all points $i$:
\begin{align}
\frac{ \partial L }{\partial w_k} = \sum_{i} \frac{ \partial L }{\partial x_i^{\ell}} \frac{ \partial x_i^\ell }{\partial w_k} 
\end{align}
where
\begin{align}
\frac{ \partial x_i^\ell }{\partial w_k} = \frac{1}{\mid \Omega_i(k) \mid} \sum_{p_j \in \Omega_i(k)} x_j^{\ell - 1}
\end{align}

Note that the above formula does not assume a specific shape for convolution kernel. Here we simply use a uniform grid kernel. In conjunction with an acceleration structure for neighbor query, e.g., grid, the convolution operator can be efficiently implemented on both CPU and GPU. In this paper, we use convolution kernels of size $3 \times 3 \times 3$. All points within each kernel cell have the same weights.

Unlike convolution in volumes, in our design, we do not use pooling. There are some advantages of doing so. First, it is no longer required to deal with point cloud downsampling and upsampling, which is not straightforward when the point attributes become high dimensional when the point cloud is processed in the network. Second, by keeping the point cloud unchanged in the entire network, acceleration structures for neighbor query only need to be built once. This significantly speeds up computation and simplifies network design. 

\paragraph{Point order.} 
A notable difference between our design and PointNet~\cite{qi-pointnet-cvpr17} is how points are ordered before being fed to the network. In PointNet, point cloud is orderless, and the training process of PointNet learns a symmetric function to turn an ordered point cloud into order invariant. However, we argue that this might not be necessary. In our method, we input points sorted in a specific order, e.g., XYZ or Morton curve~\cite{morton1966computer}, to the network and can still achieve competitive performance in the object recognition task. In this task, the order of the points only affects the final global feature vector used to predict the object category. In semantic segmentation, in principle we can leverage local features at each point, and hence point order is not necessary.

\def\atrous{\`{a}-trous }
\def\Atrous{\`{A}-trous }
\paragraph{\Atrous convolution.}
The original pointwise convolution can be easily extended to \atrous convolution by including a stride parameter that determines the gaps between kernel cells. The benefit of pointwise \atrous convolution is that it is possible to extend the kernel size, and hence the perceptive field, without actually processing too many points in the convolution. This yields significant speed up without sacrificing accuracy as to be demonstrated in our experiments.

\paragraph{Point attributes.}
For easy housekeeping in the implementation of our convolution operator, we separately store point coordinates and other point attributes such as colors, normals, or other high-dimensional features output from preceding convolutional layers. Point coordinates can be passed to any layer despite the layer depth so that they can be used for neighbor queries to determine which points can participate in the convolution at a particular point. Point attributes can then be retrieved accordingly. 



\paragraph{Relevance to geometric deep learning.}
Our pointwise convolution is relevant to geodesic convolution in geometric deep learning~\cite{bronstein-geometric-ar16}, which is more robust for tasks such as non-rigid shape correspondences and retrieval. To compute a geodesic convolution at a particular point, only neighbor points on its local surface manifold are considered. This is achieved by definition because the filter support in geodesic convolution is directly defined on the surface manifold. By contrast, our pointwise convolution operates adaptively in the 3D Euclidean space, and does not require any surface definition to operate. 

\section{Evaluations}

\paragraph{Semantic segmentation.}
We evaluate our pointwise convolutional neural network with semantic scene segmentation and object recognition. 
For scene segmentation, 
we first experiment with the S3DIS dataset~\cite{armeni_cvpr16}, which has 13 categories of indoor scene objects. Each point has 9 attributes: XYZ coordinates, RGB color, and normalized coordinates w.r.t. the room space it belongs to. To perform segmentation of a scene, each squared-meter block of the scene (measured on the floor), sampled to 4096 points, are fed into the network. The predictions of all blocks are then assembled to obtain the prediction of the entire scene. 

\begin{figure}[b]
	\def\sc{0.48}
	\def\hsp{\hspace{0.05in}}
	\def\vsp{\vspace{0.08in}}
	\centering
	\begin{subfigure}[t]{\sc\linewidth}
		\includegraphics[width=\linewidth]{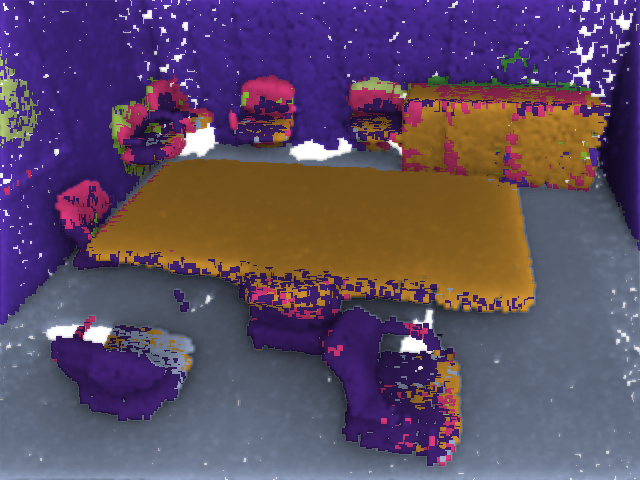}
	\end{subfigure} \hsp
	\begin{subfigure}[t]{\sc\linewidth}
		\includegraphics[width=\linewidth]{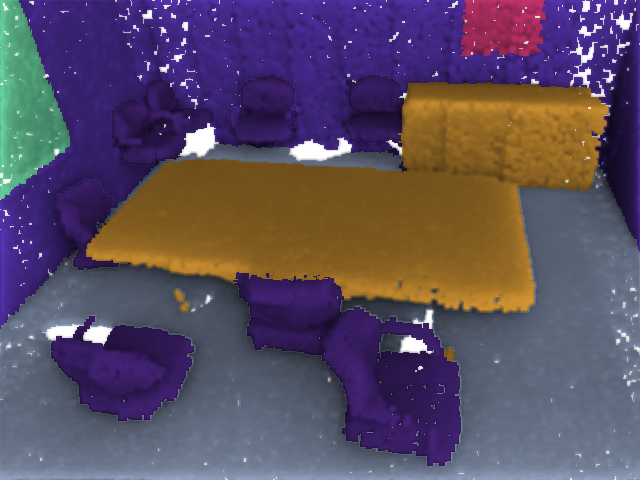}
	\end{subfigure}
	\\ \vsp
	\begin{subfigure}[t]{\sc\linewidth}
		\includegraphics[width=\linewidth]{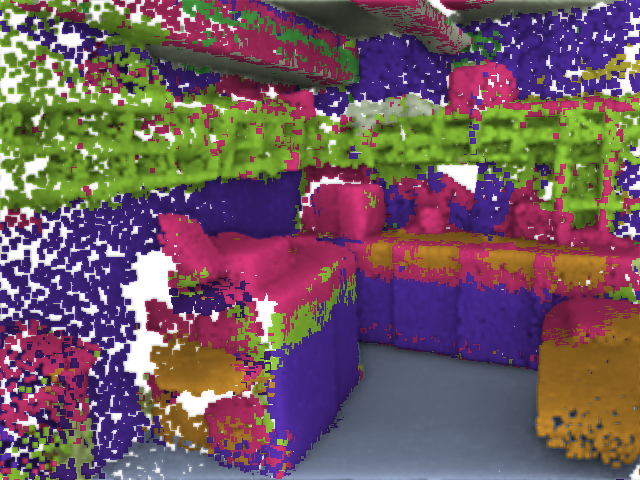}
	\end{subfigure} \hsp
	\begin{subfigure}[t]{\sc\linewidth}
		\includegraphics[width=\linewidth]{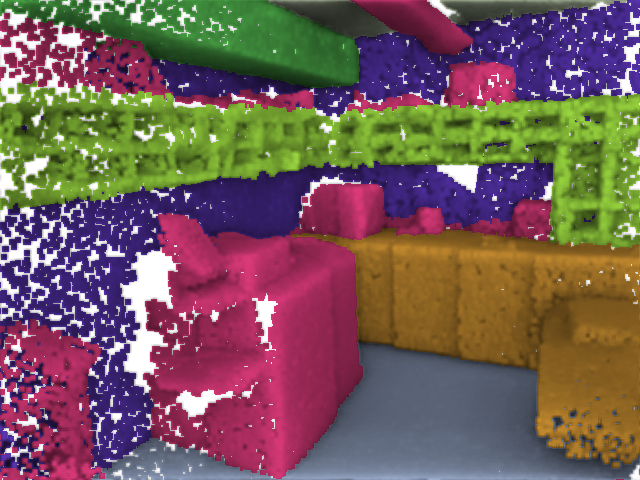}
	\end{subfigure}
	\\ \vsp
	\begin{subfigure}[t]{\sc\linewidth}
		\includegraphics[width=\linewidth]{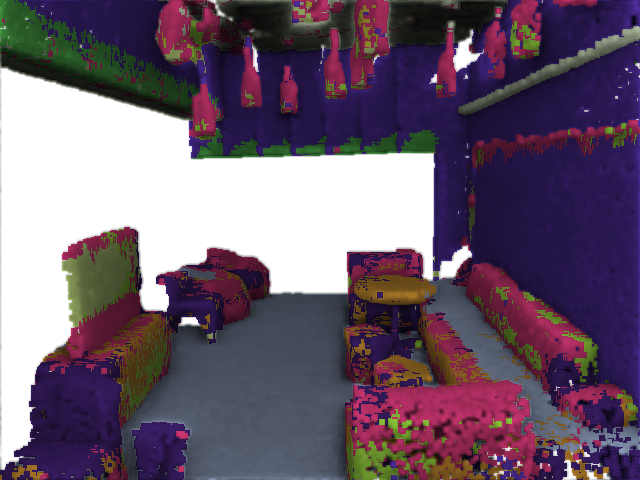}
	\end{subfigure} \hsp
	\begin{subfigure}[t]{\sc\linewidth}
		\includegraphics[width=\linewidth]{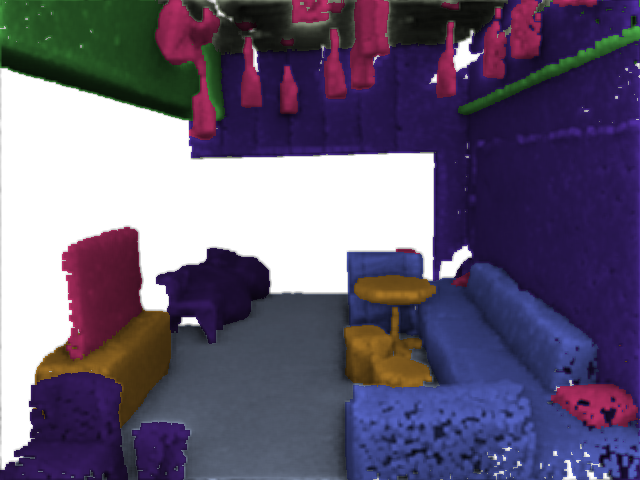}
	\end{subfigure}
	\\ \vsp
	\begin{subfigure}[t]{\sc\linewidth}
		\includegraphics[width=\linewidth]{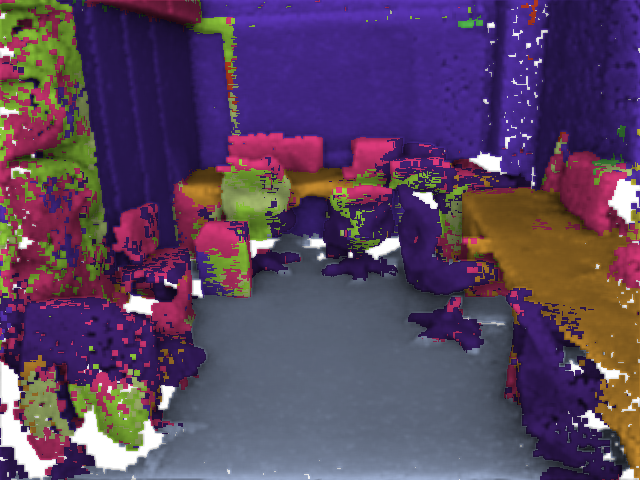}
		\caption{Our predictions}
	\end{subfigure} \hsp
	\begin{subfigure}[t]{\sc\linewidth}
		\includegraphics[width=\linewidth]{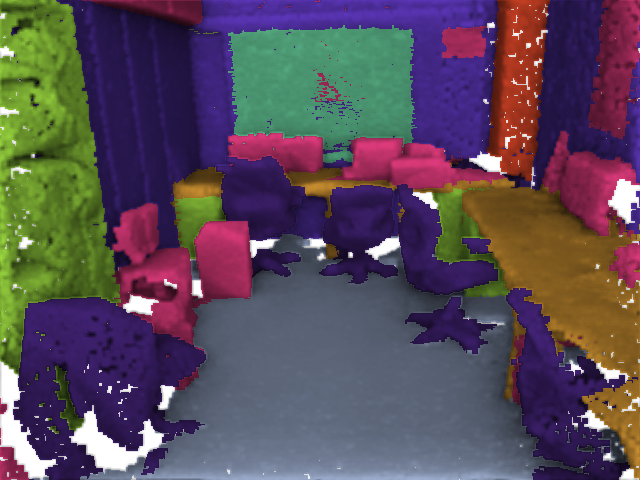}
		\caption{Ground truth}
	\end{subfigure}
	\\ \vsp
	\caption{Semantic segmentation on the S3DIS dataset~\cite{armeni_cvpr16}.}
	\label{fig:segvis}
\end{figure}

We report per-point accuracy of the semantic segmentation. As shown in Table~\ref{tab:segmentation}, our network is able to produce comparable accuracy to PointNet~\cite{qi-pointnet-cvpr17}, with the accuracy of $81.5\%$. Table~\ref{tab:s3dis_perclass} reports per-class accuracy. 
Figure~\ref{fig:segvis} shows visualization of predictions and ground truths of the scenes in the evaluation dataset.

\begin{table}[h]
	\small
	\centering
	\begin{tabular}{lll}
		\toprule
		Network & \multilinecell{Accuracy\\(per class)} & Accuracy \\
		\midrule  
		PointNet~\cite{qi-pointnet-cvpr17} & - & 87.0 \\ 
		Ours & 56.5 & 81.5 \\ 
		\bottomrule
	\end{tabular}
	\caption{Comparison of scene segmentation on S3DIS dataset~\cite{armeni_cvpr16}.}
	\label{tab:segmentation}
\end{table}

\begin{table}[t]
    \small
    \centering
    \begin{tabular}{l lllll}
        \toprule
        Network & ceiling & floor & wall & column \\ 
        \midrule  
        PointNet~\cite{qi-pointnet-cvpr17} & 98.3 & 98.8 & 83.3 & 63.4 \\ 
        Ours & 97.4 & 99.1 & 89.1 & 56.2 \\
        \midrule 
        & door & table & chair & sofa & clutter \\ 
        \midrule 
        PointNet~\cite{qi-pointnet-cvpr17} & 84.6 & 70.3 & 66.0 & 56.7 & 69.0 \\ 
        Ours & 62.9 & 73.7 & 68.4 & 54.6 & 65.2 \\
        \bottomrule
    \end{tabular}
    \caption{Per-class accuracy of semantic segmentation on S3DIS dataset~\cite{armeni_cvpr16}.}
    \label{tab:s3dis_perclass}
\end{table}

\begin{figure}[b]
	\def\sc{0.48}
	\def\hsp{\hspace{0.05in}}
	\def\vsp{\vspace{0.08in}}
	\centering
	\begin{subfigure}[t]{\sc\linewidth}
		\includegraphics[width=\linewidth]{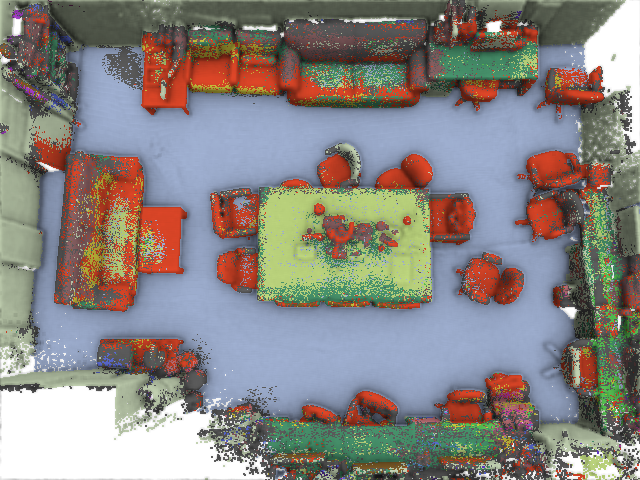}
	\end{subfigure} \hsp
	\begin{subfigure}[t]{\sc\linewidth}
		\includegraphics[width=\linewidth]{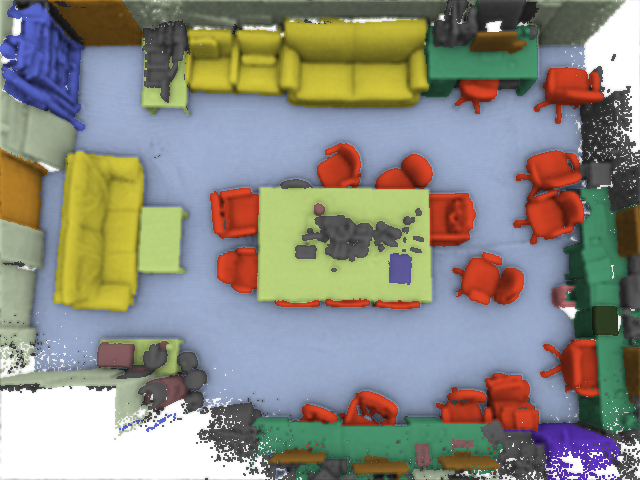}
	\end{subfigure}
	\\ \vsp
	\begin{subfigure}[t]{\sc\linewidth}
		\includegraphics[width=\linewidth]{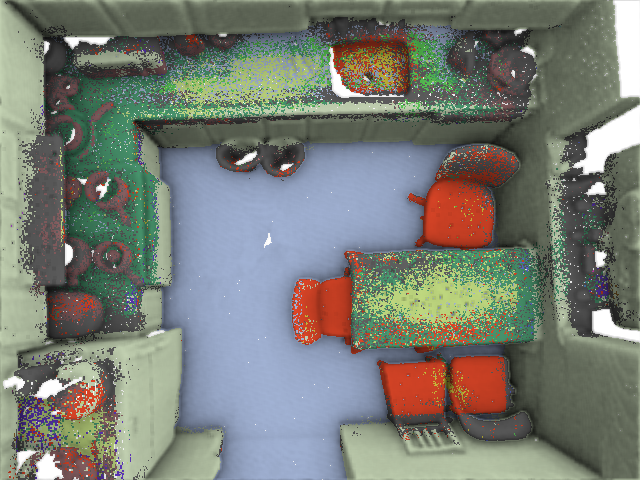}
	\end{subfigure} \hsp
	\begin{subfigure}[t]{\sc\linewidth}
		\includegraphics[width=\linewidth]{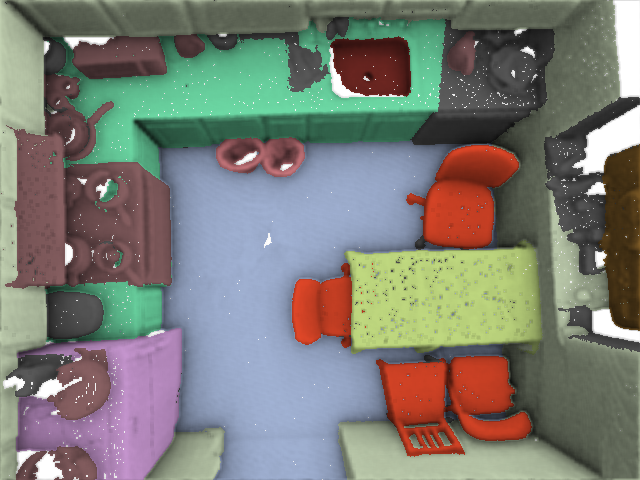}
	\end{subfigure}
	\\ \vsp
	\begin{subfigure}[t]{\sc\linewidth}
		\includegraphics[width=\linewidth]{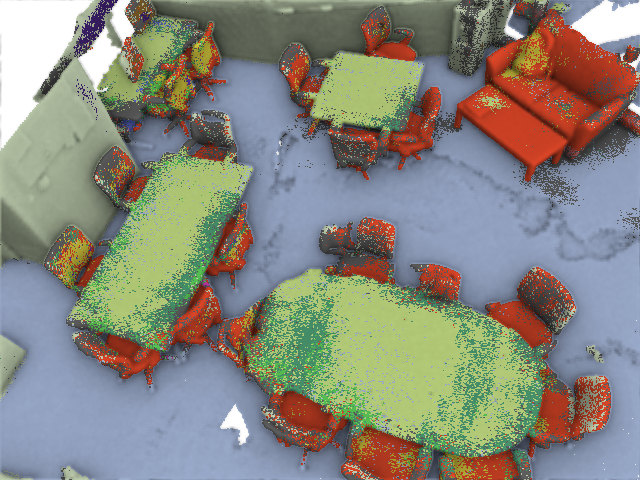}
	\end{subfigure} \hsp
	\begin{subfigure}[t]{\sc\linewidth}
		\includegraphics[width=\linewidth]{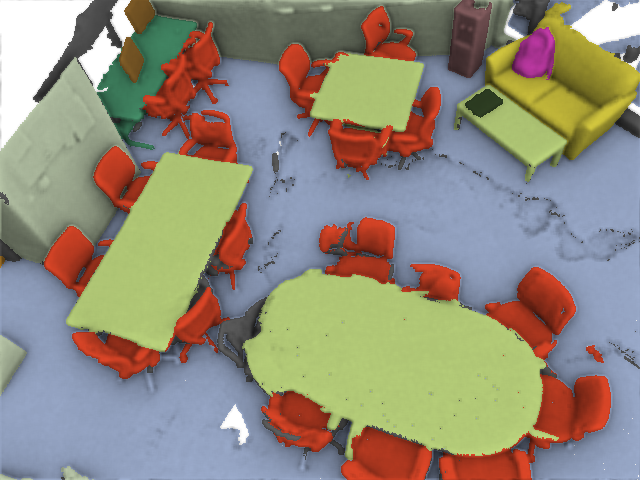}
	\end{subfigure}
	\\ \vsp
	\begin{subfigure}[t]{\sc\linewidth}
		\includegraphics[width=\linewidth]{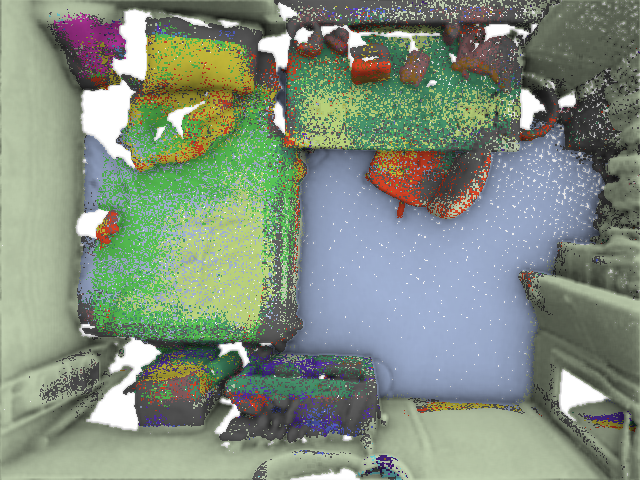}
		\caption{Our predictions}
	\end{subfigure} \hsp
	\begin{subfigure}[t]{\sc\linewidth}
		\includegraphics[width=\linewidth]{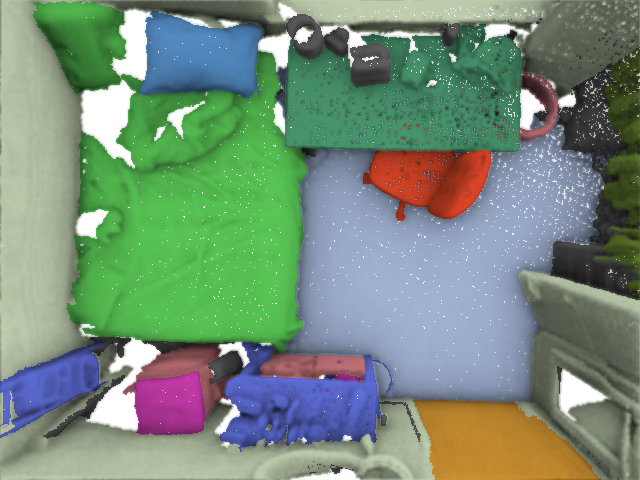}
		\caption{Ground truth}
	\end{subfigure}
	\\ \vsp
	\caption{Semantic segmentation on SceneNN dataset~\cite{hua-scenenn-3dv16}.}
	\label{fig:vis_scenenn}
\end{figure}

To further test semantic segmentation with more categories and more complex indoor scenes, we annotate 76 scenes from the SceneNN dataset~\cite{hua-scenenn-3dv16} with 40 categories defined by the NYU v2 dataset~\cite{nathan-nyu-eccv12}. Scenes in this dataset appear to be more cluttered, which poses great challenges to semantic segmentation. We use 56 scenes for training, and 20 scenes for evaluation. In each scene, a $2 \times 2$ sqm. window with stride 0.2 meter and height 2 meters is used to scan the floor area, resulting in approximately 30K scene blocks for training and 15K blocks for testing. Each block is sampled to 4096 points.

For SceneNN dataset, we additionally compare with VoxNet~\cite{maturana-voxnet-iros15}, a voxel-based representation technique, and SemanticFusion~\cite{handa-semantic-ar16}, a multi-view 2D-3D semantic segmentation with RGB-D images. 
For VoxNet~\cite{maturana-voxnet-iros15}, we apply their network to predict labels of scene blocks as described above and gather all outputs into a final scene prediction. 
For SemanticFusion~\cite{handa-semantic-ar16}, we perform 2D semantic segmentation on the RGB-D images independently and then integrate all 2D predictions to a 3D point cloud to generate the final segmentation.

The visualization of the predictions and ground truth are shown in Figure~\ref{fig:vis_scenenn}. It can be seen that structures like wall and floor have very good accuracy, and small objects are moderately well segmented. A notable issue is noise due to prediction inconsistency in the overlap regions of the blocks. This could be addressed by a conditional random field and would be an interesting future work. 

Table~\ref{tab:scenenn_perclass} reports the accuracy of a few common categories. While structures and chairs are quite accurate, table and desk are often ambiguous, resulting in lower accuracy for both classes. 
In general, the performance of VoxNet~\cite{maturana-voxnet-iros15} is inferior to ours and SemanticFusion~\cite{handa-semantic-ar16} due to limited resolution (we used $64^3$ volume). Our method works competitively to SemanticFusion, but note that our method does not apply any label smoothing while SemanticFusion has a conditional random field to remove noise after propagating predictions from 2D to 3D. 

\begin{table}[h]
    \small
    \centering
    \begin{tabular}{l lllll}
        \toprule
        Network & wall & floor & chair & table & desk \\ 
        \midrule  
        VoxNet~\cite{maturana-voxnet-iros15}      & 82.8 & 74.3 & 3.1  & 0.8  & 5.4 \\
        SemanticFusion~\cite{handa-semantic-ar16} & 72.8 & {94.4} & 46.3 & {70.1} & 28.1 \\ 
        Ours                                      & {93.8} & 88.6 & {58.6} & 23.5 & {29.5} \\       
        \bottomrule
    \end{tabular}
    \caption{Per-class accuracy of semantic segmentation on SceneNN dataset~\cite{hua-scenenn-3dv16}.}
    \label{tab:scenenn_perclass}
\end{table}

\paragraph{Object recognition.}
We evaluate object recognition with two datasets, ModelNet40~\cite{wu-3dshapenets-cvpr15} and ObjectNN~\cite{huashrec}. 
ModelNet40 is a CAD model dataset of 40 categories which has served as a standard benchmark for object recognition in the recent years. On the other hand, ObjectNN is an object dataset from RGB-D scene reconstruction mixed with CAD models for studying 3D object retrieval. 
Objects in ObjectNN is particularly difficult to classify because they are reconstructed from noisy RGB-D data and often has missing parts. 

For object recognition, our point attributes are simple XYZ coordinates. In fact, we also trained the network with point attributes set to one, making the convolution equivalent to density estimation, and found no significant change in accuracy.
Our results on ModelNet40 are shown in Table~\ref{tab:modelnet40}.
\begin{table}[h]
 \small
\renewcommand{\arraystretch}{1}
\centering
\begin{tabular}{llllll}
    \toprule
    Network & \multilinecell{Accuracy\\(per class)} & Accuracy\\
    \midrule  
    VoxNet~\cite{maturana-voxnet-iros15}       &  83       & -        \\
    MO-SubvolumeSup~\cite{qi-volumetric-cvpr16}                    &  86       & 89.2     \\
    PointNet~\cite{qi-pointnet-cvpr17}           &  86.2     & 89.2     \\
    PointNet++~\cite{qi2017pointnetplusplus}                         &  -        & 90.7     \\
    Ours                                       &  81.4        & 86.1     \\
    \bottomrule
\end{tabular}
\caption{Comparison of performance of network architectures using 3D object representations on the ModelNet40 dataset~\cite{wu-3dshapenets-cvpr15}.}
\label{tab:modelnet40}
\end{table}
As can be seen, our network performs comparably to state-of-the-art methods. Note that compared to VoxNet~\cite{maturana-voxnet-iros15}, our input point cloud is more compact. Our network is also significantly simpler in design compared to PointNet~\cite{qi-pointnet-cvpr17} and PointNet++~\cite{qi2017pointnetplusplus} while being close to their accuracy.

The results on ObjectNN are shown in Table~\ref{tab:objectnn}.
\begin{table}[b]
\small
\renewcommand{\arraystretch}{1}
\centering
\begin{tabular}{lll}
	\toprule
	Network & \multilinecell{Accuracy\\(per class)} & Accuracy \\
	\midrule  
	PointNet~\cite{qi-pointnet-cvpr17} & 57.1 & 65.6 \\
	Ours                             & 57.1 & 65.1 \\
	\bottomrule
\end{tabular}
\caption{Comparison of object recognition accuracy on the ObjectNN dataset~\cite{huashrec}.}
\label{tab:objectnn}
\end{table}
In this dataset, again our method performs comparably to PointNet, but overall both methods are less effective due to the ambiguity in learning features from both CAD models and RGB-D objects. 

Table~\ref{tab:modelnet40_perclass} and Table~\ref{tab:objectnn_perclass} further provide per-class accuracy on the ModelNet40 and the ObjectNN dataset, respectively.

\begin{figure}[t]
	\def\sc{0.495}
	\begin{subfigure}[t]{\sc\linewidth}
		\includegraphics[width=\linewidth]{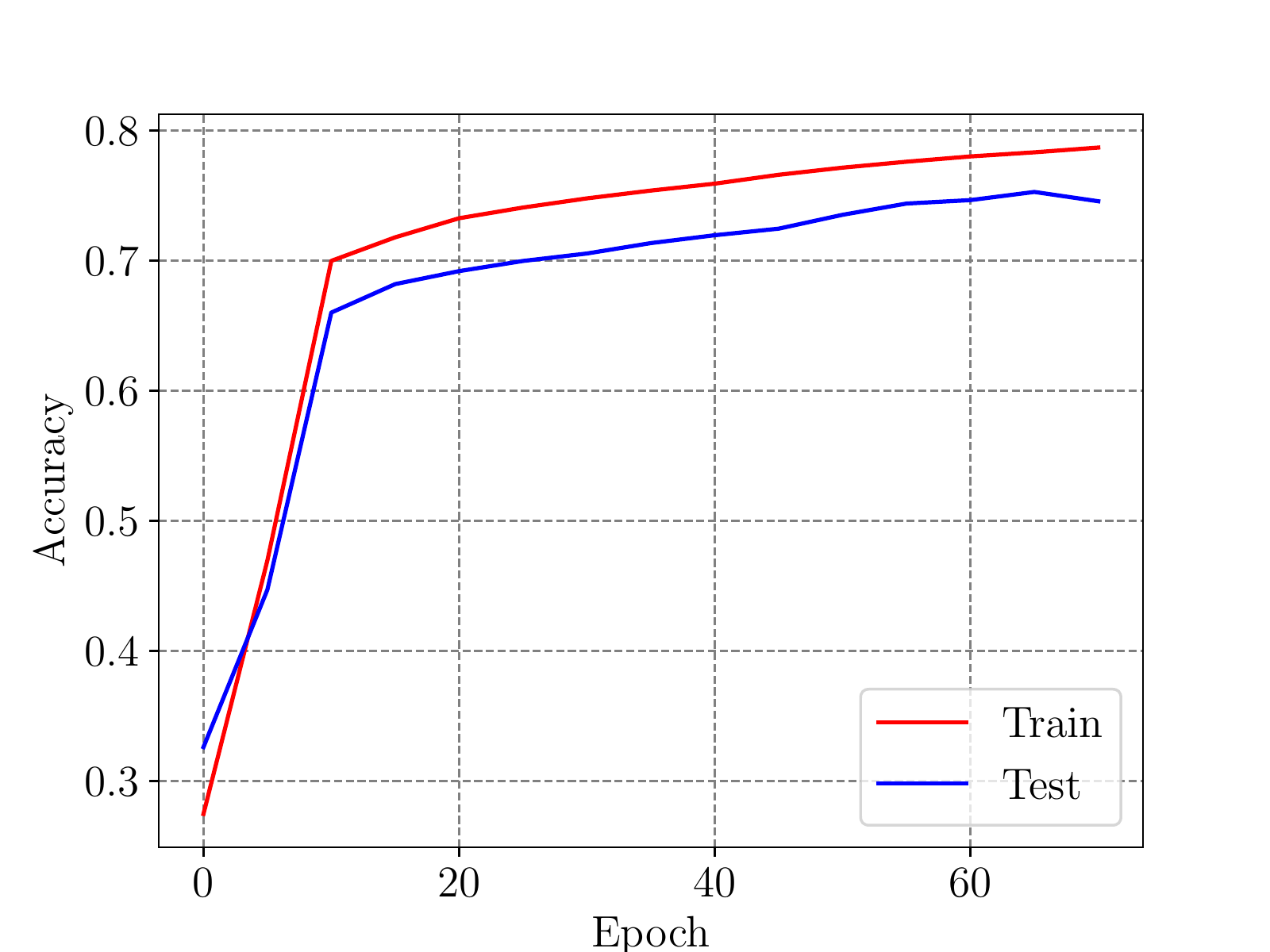}
		\caption{Scene segmentation}
	\end{subfigure}
	\begin{subfigure}[t]{\sc\linewidth}
		\includegraphics[width=\linewidth]{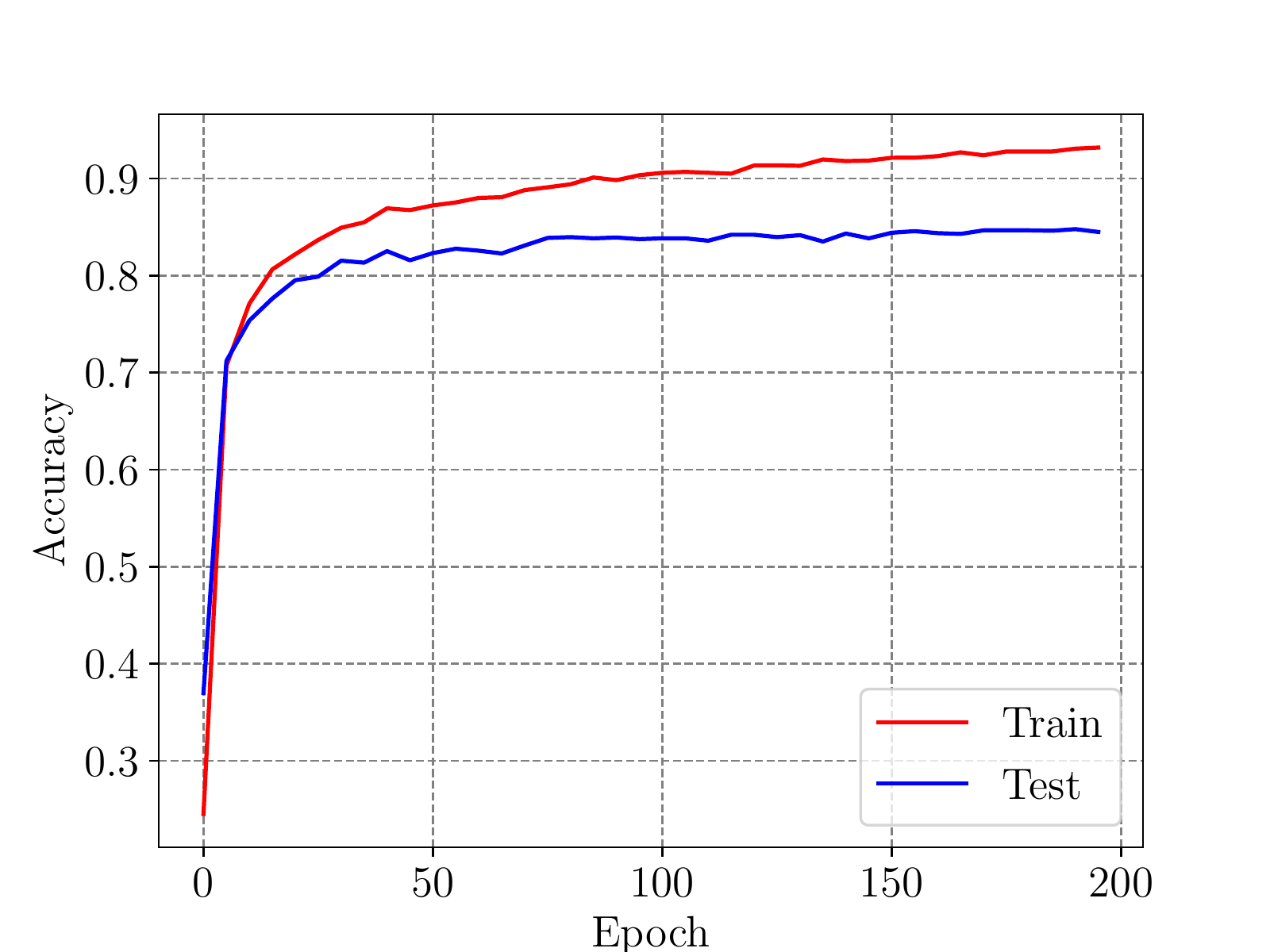}
		\caption{Object recognition}
	\end{subfigure}
	\caption{Train and test accuracy over time.}
	\label{fig:traintest}
\end{figure}

\paragraph{Convergence.} Figure~\ref{fig:traintest} shows a plot of the training and test accuracy of our networks over time. The graph shows that our pointwise convolutional neural network can be trained effectively. 

\paragraph{Ablation experiments.}
Here we analyze the effectiveness of pointwise convolution. We first start with with a basic 4-layer model as in Figure~\ref{fig:network}. The accuracy improvement when more features are added are presented in Table~\ref{tab:ablation}. As can be seen, feature concatenation, \atrous convolution, SELU activation, and dropout each contributes a small improvement to the final result. 

\begin{table}[h]
    \small
	\centering
	\begin{tabular}{llllll}
		\toprule
		Base & Concat. & \Atrous & SELU & Dropout & Accuracy \\
		\midrule
		\checkmark &  &  &  &  & 78.6 \\
		\checkmark & \checkmark & & & & 78.0 \\	
		\checkmark & & \checkmark & & & 75.0 \\
		\checkmark & \checkmark & \checkmark &  &  & 82.5 \\
		\checkmark & &  &  \checkmark &  & 81.7 \\	
		\checkmark & \checkmark &  &  \checkmark &  & 81.9 \\		
		\checkmark & \checkmark & & \checkmark & \checkmark & 85.2 \\		
		\checkmark & \checkmark & \checkmark & \checkmark & \checkmark & 86.1 \\
		\bottomrule
	\end{tabular}
	\caption{Ablation experiment. Accuracy improvement is achieved when pointwise convolution is combined with feature concatenation (Concat.), \atrous convolution, self-normalization activation function (SELU), and dropout.}
	\label{tab:ablation}
\end{table}

\paragraph{Point order.} 
In object recognition, the order of the input points determine the orders of the features in the fully connected layers. As long as this layer has an order, it is sufficient to discriminate their features and predict the categories. 
We experiment with different orders of the input point set and report the results in Table~\ref{tab:orders_adaptive}(a). We found that point orders sorted by space filling curve techniques such as Morton curve~\cite{morton1966computer} yields comparable accuracy, which means that it is sufficient to just follow an order, but not a particular one. However, a benefit is that space filling curves organize points such that nearby points in space are stored close to each other in memory, allowing more memory coherence.

\paragraph{Neighborhood radius.}
So far we have been setting the radius for neighbor query as constant in each convolution layer. In our experience, this works well for both tasks. We also explore the capability of adaptive radius using k-nearest neighbors. The modification for the convolution operator is as follows. 

At each point, a k-nearest neighbor is performed, and the query radius is set to the distance to the furthest neighbor. This radius is used each time neighbor points have to be queried for convolution.
To compute gradients for backpropagation for this operator, it is worth noting that in this case, neighbor lookup is no longer symmetric. Therefore, at a point $j$, it is required to look up all points $i$ such that point $i$ can contribute to point $j$ in the forward convolution. 

We compare the performance of the k-nearest neighbor and the fixed radius convolution for object recognition task. The result is shown in Table~\ref{tab:orders_adaptive}(b). In general, we found no significant difference in terms of accuracy. 
 
\begin{table}[h]
    \small
	\centering
	\begin{minipage}[t]{0.4\linewidth}
		\centering
		\begin{tabular}{ll}
			\toprule
			Order & Accuracy \\
			\midrule  
			ZYX  & 86.1 \\
			Morton & 86.0 \\
			\bottomrule
		\end{tabular}
		\centerline{(a)}
	\end{minipage}
	\begin{minipage}[t]{0.59\linewidth}
		\begin{tabular}{ll}
			\toprule
			Neighbor query & Accuracy \\
			\midrule  
			Fixed-size radius  & 86.1 \\
			K-nearest neighbor & 85.7 \\
			\bottomrule
		\end{tabular}
		\centerline{(b)}
	\end{minipage}
	\caption{(a) Object recognition with different ways of ordering the input point cloud. (b) Object recognition with convolution using neighbor queries with adaptive radius.}
	\label{tab:orders_adaptive}
\end{table}


\paragraph{Deeper networks.} 
Finally, we study the capability of learning with deeper networks using pointwise convolution. From the basic model, we increase the number of layers from 4 to 8 and 16, and then retrain from scratch. The performance are reported in Table~\ref{tab:deepconv} below. 
\begin{table}[b]
    \small
	\centering
	\begin{tabular}{ll}
		\toprule
		Network & Accuracy \\
		\midrule  
		4 layers & 86.1 \\
		8 layers & 82.1 \\
		16 layers & 82.6 \\
		\bottomrule
	\end{tabular}
	\caption{Deep pointwise convolutional neural network. We compare object recognition performance with 4-, 8-, and 16-layer architecture.}
	\label{tab:deepconv}
\end{table}
Generally, it takes longer to train networks with 8 and 16 layers, resulting in slightly slower accuracy. Experimenting the training with residual learning~\cite{he-resnet-cvpr16} would be an interesting future work.


\paragraph{Running time.}
A key challenge when implementing pointwise convolution is how to perform fast nearest neighbor query without impacting too much the network training and prediction time. To make the training feasible, we choose to use a grid for neighbor query because this is a lightweight and GPU-friendly data structure to build and query \emph{on the fly}. In fact, we experimented with kd-tree, but found that on modern CPUs and GPUs, a kd-tree query does not outperform a grid unless the number of points are more than 16K points, not to mention extra time needed for tree construction that has $O(n \log n)$ complexity. 

Our pointwise convolution is currently implemented with Tensorflow. We report the running time, including grid build and query each time convolution is invoked, as follows. For a batch size of 128 point clouds, each with 2048 points, a forward convolution of our network takes 1.272 seconds on an Intel Core i7 6900K with 16 threads, and a backward propagation takes 2.423 seconds to compute the gradients. Our GPU implementation on an NVIDIA TITAN X can further improve the running time for about 10\%. 
Compared to PointNet~\cite{qi-pointnet-cvpr17} and VoxNet~\cite{maturana-voxnet-iros15} which leverage Tensorflow's optimized convolution operators, our pointwise convolution is not yet engineering optimized. Our training time is about $2\times$ slower which we currently compensate by using multiple CPUs and GPUs.

\begin{table*}[t]
    \small
    \centering
    \begin{tabular}{l lllll lllll}
        \toprule
        Network                             & airplane & bathtub & bed  & bench & bookshelf & bottle & bowl  & car  & chair & cone  \\ 
        \midrule  
        PointNet~\cite{qi-pointnet-cvpr17}  & 100      & 80.0    & 94.0 & 75.0  & 93.0      & 94.0   & 100.0 & 97.9 & 96.0  & 100.0 \\ 
        Ours                                & 100      & 82.0    & 93.0 & 68.4  & 91.8      & 93.9   & 95.0  & 95.6 & 96.0  & 80.0  \\ 
        \midrule 
        & cup   & curtain & desk & door & dresser & flower pot & glass box & guitar & keyboard & lamp \\ 
        \midrule  
        PointNet~\cite{qi-pointnet-cvpr17}  & 70.0  & 90.0    & 79.0 & 95.0 & 65.1    & 30.0       & 94.0      & 100.0  & 100.0    & 90.0 \\ 
        Ours                                & 60.0  & 80.0    & 76.7 & 75.0 & 67.4    & 10.0       & 80.8      & 98.0   & 100.0    & 83.3 \\
        \midrule 
        & laptop & mantel & monitor & night stand & person & piano & plant & radio & range hood & sink \\
        \midrule 
        PointNet~\cite{qi-pointnet-cvpr17}  & 100.0  & 96.0   & 95.0    & 82.6        & 85.0   & 88.8  & 73.0  & 70.0  & 91.0       & 80.0 \\ 
        Ours                                & 95.0   & 93.9   & 92.9    & 70.2        & 89.5   & 84.5  & 78.8  & 65.0  & 88.9       & 65.0 \\
        \midrule 
        & sofa & stairs & stool & table & tent & toilet & tv stand & vase & wardrobe & xbox \\
        \midrule 
        PointNet~\cite{qi-pointnet-cvpr17}  & 96.0 & 85.0   & 90.0  & 88.0  & 95.0 & 99.0   & 87.0     & 78.8 & 60.0     & 70.0\\ 
        Ours                                & 96.0 & 80.0   & 83.3  & 90.9  & 90.0 & 94.9   & 84.5     & 81.3 & 30.0     & 75.0 \\
        \bottomrule
    \end{tabular}
    \caption{Per-class accuracy of object recognition on the ModelNet40 dataset. Average: PointNet: 86.3. Ours 81.4.}
    \label{tab:modelnet40_perclass}
\end{table*}

\begin{table*}[t]
	\small
    \centering
    \begin{tabular}{l lllll lllll}
        \toprule
        Network                               & chair & display & desk & book & storage & box   & table & bin  & bag  & keyboard \\ 
        \midrule  
        PointNet~\cite{qi-pointnet-cvpr17}    & 84.2  & 85.4    & 56.7 & 30.1 & 62.5    & 23.8  & 80.0  & 75.0 & 47.4 & 82.4     \\ 
        Ours                                  & 83.1  & 85.4    & 70.0 & 57.7 & 45.8    & 23.8  & 60.0  & 65.0 & 36.8 & 88.2     \\
        \midrule 
        & sofa & bookshelf & pillow & machine & pc case & light & oven & cup  & printer & bed \\ 
        \midrule 
        PointNet~\cite{qi-pointnet-cvpr17}    & 76.5 & 23.1      & 84.6   & 18.2    & 36.4    & 77.8  & 60.0 & 37.5 & 50.0    & 28.6\\ 
        Ours                                  & 88.2 & 38.5      & 76.9   & 18.2    & 54.5    & 88.9  & 30.0 & 75.0 & 12.5    & 42.9\\
        \bottomrule
    \end{tabular}
    \caption{Per-class accuracy of object recognition on the ObjectNN dataset. 
        Average: PointNet: 56.0. Ours: 57.1.}
    \label{tab:objectnn_perclass}
\end{table*}

\section{Conclusion}
In this paper, we proposed pointwise convolution and leveraged it to build convolutional neural networks for scene understanding with point cloud data. We demonstrated two scene understanding applications including scene segmentation and object recognition. 
We showed that it is practical to simply sort input point clouds in a specific order before feature learning for object classification. Our pointwise convolution can offer competitive accuracy while being simple to implement, allowing us to create effective and simple neural networks for learning local features of point clouds. 

There are several research avenues to be further explored. For example, finding a robust solution to handle large-scale point clouds for scene understanding would be an interesting future work. Currently, we just circumvent the large-scale issue in semantic segmentation by simply dividing the scene into blocks and resample each block to fixed number of points for prediction. 
In addition, it would be of great interest to extend pointwise convolutional neural networks to geometry point cloud processing~\cite{yu-pointupsampling-cvpr18}, or explore the connection of pointwise convolution to tensor voting~\cite{wu-tensor-pami12}, which was used in the literature to detect structures in a local point neighborhood. 

\appendix 
\section{Layer Visualization}
Intuitively, pointwise convolution works by summarizing local spatial point distributions to build feature vectors for each point in a point cloud. As shown in per-class accuracy tables, local features work the most effectively in classifying structures such as ceiling, floor, or walls and common furniture such as tables and chairs. In our observation, it is quite challenging to differentiate between tables (for dining) and desks (for study and work). 

\begin{figure}[h]
    \def\sc{0.23}
    \def\scb{0.35}
    \def\hsp{\hspace{0.05in}}
    \begin{subfigure}[t]{\sc\linewidth}
        \centering
        \includegraphics[width=\scb\linewidth]{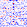}
        \caption{Layer 1}
    \end{subfigure} \hsp
    \begin{subfigure}[t]{\sc\linewidth}
        \centering
        \includegraphics[width=\scb\linewidth]{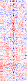}
        \caption{Layer 2}
    \end{subfigure} \hsp
    \begin{subfigure}[t]{\sc\linewidth}
        \centering
        \includegraphics[width=\scb\linewidth]{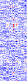}
        \caption{Layer 3}
    \end{subfigure} \hsp
    \begin{subfigure}[t]{\sc\linewidth}
        \centering
        \includegraphics[width=\scb\linewidth]{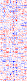}
        \caption{Layer 4}
    \end{subfigure}
    \caption{Visualization of the filters in pointwise convolution network for object recognition.}
    \label{fig:vis_filter}
\end{figure}

We visualize the filters of the first four layers in the object recognition network in Figure~\ref{fig:vis_filter}. 
Here we display each $3 \times 3 \times 3$ filter on a row in the visualization. The number of rows is equal to the product of the total number of input and output channels of each filter (27 for the first layer, and 81 for the subsequent layers).  
In the visualization, blue and red represent positive and negative values, respectively. White represents zero. This shows that the filters in the network are relatively sparse and smooth. We also observed that positive and negative values dominate the filters interchangeably in each layer.

\newpage
\paragraph*{Acknowledgement.}
We thank Quang-Hieu Pham for helping with the 2D-to-3D semantic segmentation experiment and proofreading the paper, Quang-Trung Truong and Benjamin Kang Yue Sheng for their kind support for the neural network training experiments.

Binh-Son Hua and Sai-Kit Yeung are supported by the SUTD Digital Manufacturing
and Design Centre which is supported by the Singapore National Research
Foundation (NRF). Sai-Kit Yeung is also supported by Singapore MOE Academic
Research Fund MOE2016-T2-2-154, Heritage Research Grant of the National Heritage
Board, Singapore, and Singapore NRF under its IDM Futures Funding Initiative and
Virtual Singapore Award No. NRF2015VSGAA3DCM001-014.

{\small
\bibliographystyle{ieee}
\bibliography{ss3d}
}

\end{document}